%% file: main.tex
\documentclass{article} 
\usepackage{iclr2026_conference,times}

\input{math_commands.tex}

\usepackage{hyperref}
\usepackage{url}
\usepackage{tabularx}
\usepackage{booktabs} 
\usepackage{array}
\usepackage{multirow}
\usepackage{makecell}
\usepackage{amsmath}
\usepackage{graphicx}
\usepackage{verbatim}
\usepackage[most]{tcolorbox}

\usepackage{algorithm}
\usepackage{algpseudocode}
\usepackage{bm} 

\usepackage[table]{xcolor}  
\definecolor{first}{RGB}{190, 225, 200}   
\definecolor{second}{RGB}{230, 240, 185}  
\definecolor{third}{RGB}{255, 250, 195}   

\title{Retrieval, Refinement, and Ranking for Text-to-Video Generation via Prompt Optimization and Test-Time Scaling}

\author{Zillur Rahman
\thanks{Correspondance to: zillur.mle@gmail.com} \\
Algoverse AI\\
\And
Alex Sheng \\
Algoverse AI \\
\And
Cristian Meo \\
Algoverse AI\\
}

\iclrfinalcopy 
\begin{document}

\maketitle

\begin{abstract}
While large-scale datasets have driven significant progress in Text-to-Video (T2V) generative models, these models remain highly sensitive to input prompts, demonstrating that prompt design is critical to generation quality. Current methods for improving video output often fall short: they either depend on complex, post-editing models, risking the introduction of artifacts, or require expensive fine-tuning of the core generator, which severely limits both scalability and accessibility. In this work, we introduce 3R, a novel RAG based prompt optimization framework. 3R utilizes the power of current state-of-the-art T2V diffusion model and vision language model. It can be used with any T2V model without any kind of model training. The framework leverages three key strategies: RAG-based modifiers extraction for enriched contextual grounding, diffusion-based Preference Optimization for aligning outputs with human preferences, and temporal frame interpolation for producing temporally consistent visual contents. Together, these components enable more accurate, efficient, and contextually aligned text-to-video generation. Experimental results demonstrate the efficacy of 3R in enhancing the static fidelity and dynamic coherence of generated videos, underscoring the importance of optimizing user prompts.
\end{abstract}
\section{Introduction}
Due to exciting advancements in diffusion-based generative models and large-scale training procedures, modern text-to-video (T2V) models have achieved impressive capabilities for using natural language prompts to generate photorealistic video content \citep{peebles2023scalablediffusionmodelstransformers} \citep{ramesh2022hierarchicaltextconditionalimagegeneration}. 


Despite rapid uptake in natural language processing (NLP) and subsequent innovations in text-to-image (T2I) generation \cite{podell2023sdxlimprovinglatentdiffusion, esser2024scalingrectifiedflowtransformers}, and improving image aesthetics \citep{chen2024catcatnotdog}, their impact on video quality is limited \citep{hao2023optimizingpromptstexttoimagegeneration}. Besides, the application of test-time optimization \citep{zhang2025surveytesttimescalinglarge} in text-to-video settings remains in early exploratory stages, with meaningful open challenges \citep{gu2025ccc}. This presents valuable opportunities to address problems in T2V like prompt adherence, visual quality, physical plausibility, and temporal coherence.


To generate videos from texts, users provide a short text prompt to the video generation model. Recent works show that a long detailed text prompt generates better quality videos than the short user provided prompt \citep{hao2023optimizingpromptstexttoimagegeneration, yang2025cogvideoxtexttovideodiffusionmodels}. This underscores the importance of enhancing the user prompt before feeding it to a T2V model. The short user prompts do not contain detailed contextual information required to generate vivid visual content. Moreover, videos generated from the same prompt differ in quality due to the stochastic nature of the diffusion models. So generating multiple videos from one prompt and selecting the one that better fits the user prompt with better visual quality could be effective.

To address these, in this paper, we explore avenues combining retrieval, refinement, and ranking within this emerging paradigm of inference-time compute algorithms to improve video generation quality in T2V settings. We study a black-box problem definition that is designed for direct plug-and-play applicability to off-the-shelf T2V models in real-world settings.

Our contributions can be summarized as follows:
\begin{itemize}
    \item We propose Retrieval-Refinement-Ranking (3R), a retrieval based training free prompt optimization framework for T2V generation.
    \item We propose an initial prompt refinement module that creates a detailed context rich description aligning with the user prompt.
    \item We validate the effectiveness of 3R on EvalCrafter benchmark where it achieves SOTA results among open-source models.
\end{itemize}

\section{Related Works}
\label{literature}
\textbf{Text-to-Video Models} Text-to-video generation models \citep{openai2024sora, rombach2021stablediffusion, wang2023lavie, zhang2025show1marryingpixellatent} have seen rapid advances in both model capabilities and practical accessibility. T2V models receive input prompts consisting of natural language descriptions, and comprehend described scenes, actions, objects and generate visual contents. T2V models are being used in generating animations \citep{chen2023seineshorttolongvideodiffusion}, movies \citep{zhao2025moviedreamerhierarchicalgenerationcoherent}, commercials, etc.

\textbf{Prompt Optimization Frameworks} IPO \citep{yang2025ipoiterativepreferenceoptimization} introduces an iterative optimization algorithm to align video foundation models with human preferences. It creates a human preference dataset and trains a critique model with that dataset. An iterative optimization loop is used to align a base T2V model with human preference, and thus improving subject consistency, motion smoothness, and aesthetic quality. CCC \citep{gu2025ccc} introduces a simple vision language model for text to video generation. Each candidate video is queried multiple times to get a list of issues and a content score is computed from the number of common issues. Based on those issues, initial prompts are refined to generate better results. In \citep{Gao_2025_CVPR}, authors introduced RAPO: a RAG based prompt optimization model for text to video generation. A dataset is used to extract relevant modifiers to augment the user prompt. Then a fine-tuned Llama3 model \citep{grattafiori2024llama} is used to refactor the augmented prompt into the format of training prompts. Finally, another fine-tuned LLama3 is used to select the better prompt between the refactored prompt and a refined user prompt. Google publishes VISTA \citep{long2025vistatesttimeselfimprovingvideo}, one of the most computationally expensive models where pairwise video comparison is used to select candidate videos that are evaluated by multi-modal language models for their visual, audio, and context quality. Then, LLMs review the issues and refine the original prompts to generate videos again. The entire model runs for maximum 5 iterations and each iteration can have maximum 30 videos, making it 150 videos per prompt.

\section{Method}
This section discusses the key design choices of 3R. The overall pipeline is illustrated in Fig. \ref{fig:model} and a pseudo algorithm is illustrated in Appendix \ref{app:algorithm}.
\begin{figure}[t]
    \centering
    \includegraphics[width=1.0\linewidth]{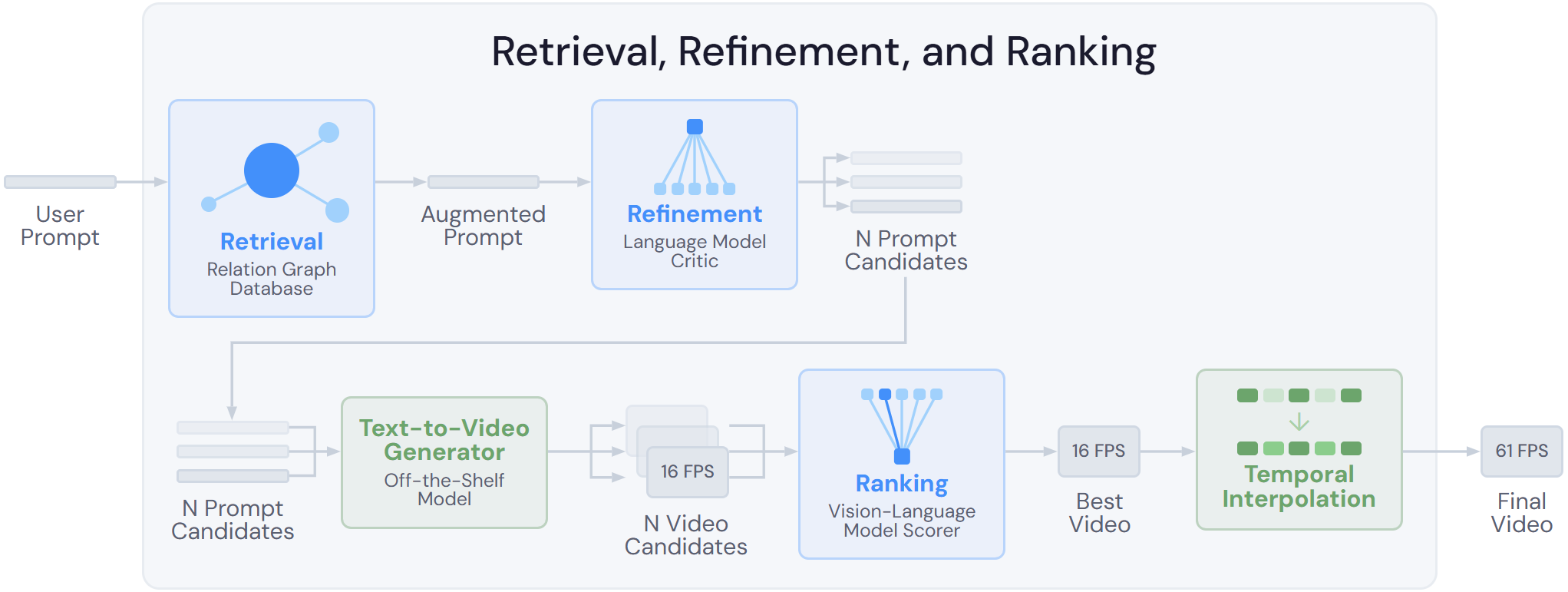}
    \caption{\textbf{Overview of 3R pipeline.} A short user prompt $I$ is used to extract a few relevant subject, scene, actions modifiers from a relation database $\mathcal{D}$. Then ${M}_{LLM}$ is used to merge those modifiers iteratively to the original user prompt to get detaild prompt $P_m$, and ${R}_{LLM}$ checks $P_m$ for any contradictory or missing information from the original prompt $I$, and generate $N$ refined prompts. The refined prompts are fed to a T2V base model $\mathcal{G}$ to generate initial videos for each prompt. Next, a video selection model selects the best candidate based on a question answering test, and a temporal interpolation network enhances temporal consistency of the final video.}
  
    \label{fig:model}
\end{figure}

\paragraph{Modifiers}
Using synthetic data augmentation to rearrange knowledge for more data-efficient learning has been proven an effective pathway to mitigate the challenge of adapting a pre-trained model to a small corpus of domain-specific documents \citep{entigraph}. Its main purpose is to overcome the model’s context limitations, enabling effective context construction for diverse user queries. Given an original user prompt intent $I$, first, scene modifiers $p_j$ are extracted from a relation database $\mathcal{D}$ using a pre-trained sentence transformer. Then using cosine similarity score, we select scenes from the relation graph if it is above a threshold $\tau$. Each scene comes with its list of subject, action and environment modifiers. We select top-k modifiers for each scene. 
\begin{equation}
P_{ret} = \{ p_j \in \mathcal{D} \mid \text{sim}(\phi(I), \phi(p_j)) > \tau \}  
\end{equation}  
We iteratively merge each modifier with $I$ using a pre-trained LLM $M_{LLM}$ in a few-shot manner. Existing method like RAPO \citep{Gao_2025_CVPR} uses a large comma separated list of all the modifiers as an initial description and prompts the LLM to merge each modifier. However, this process sometimes generates incorrect and misleading results since some modifiers have little to no relevance to the $I$. To mitigate this issue, we initialize the description with only $I$. We show such an example in Appendix Table \ref{tab:merge}.
\begin{equation}
P_{m} = M_{LLM} (I \mid P_{ret})
\end{equation}  

\paragraph{Refine Descriptions}
After merging is completed, we get a detailed description $P_m$ of each user prompt. To eliminate any contradictory, misleading information or add any useful missing information, we prompt an LLM to further refine the description. This LLM aims to refine $P_m$ based on information such as characters, actions, attributes like color, counts from $I$. This step is crucial for quality video generation. In our experiment section, we demonstrate the importance of the initial prompt. If the information in the initial prompt is not coherent, the generated video will not represent user intents. We use $R_{LLM}$ in a few-shot manner and generate $N$ distinct detailed prompts, maintaining the original user intent. The goal is to generate multiple videos that may have different positive and negative aspects so that we can choose the best candidate as the final video. The prompt for this step is illustrated in Appendix \ref{app:refinement}.
\begin{equation}
\{P_n\}_{n=1}^N = R_{LLM} (I \mid P_{m}) 
\end{equation} 

\paragraph{Video Generation}
Each prompt in our approach is passed to a T2V base model $\mathcal{G}$. T2V model is treated as a black box that is only assumed to take a natural language input prompt and return a generated video output. This setup preserves practical applicability, as our approach does not require access beyond black-box text-to-video queries. By adhering to this framing, our algorithm is applicable to any off-the-shelf T2V model, regardless of whether they are open-source models or proprietary inference APIs.
\begin{equation}
    \{V_n\}_{n=1}^N = \{ \mathcal{G}(P_n) \mid P_n \in \{P_1, \dots, P_N\} \}
\end{equation}
We explain the video selection and enhancement sections in details in Appendix \ref{app:method}.

\section{Experiments}
\textbf{Experimental Setup}
We use the EvalCrafter \citep{liu2023evalcrafter} benchmark for quantitative performance evaluation. This comprehensive text2video evaluation benchmark has 17 raw dimensions such as clip score, motion score, face consistency score, etc. These raw metrics are aggregated into 4 categories: Text-Video Alignment, Visual Quality, Motion Quality and Temporal Consistency.
We compare our approach to 4 other models that reported their performance on EvalCrafter benchmark: Lavie \citep{wang2023lavie} with original short prompts, IPO \citep{yang2025ipoiterativepreferenceoptimization}, Show-1 \citep{zhang2025show1marryingpixellatent} and Videocrafter2 \citep{chen2024videocrafter2overcomingdatalimitations}. We reproduced the IPO results, and for others, we use the results and metrics reported in the EvalCrafter benchmark leader-board. We report the implementation details in Appendix \ref{app:implementation}.

\subsection{Results}
Table \ref{tab:evalcrafter} reports the results of 3R in comparison with Show-1, LaVie, IPO, and Videocrafter2 on the four benchmark metrics of EvalCrafter. 3R approach achieves the highest total score, demonstrating the effectiveness of our inference-time approach for improving text-to-video performance. Compared with the LaVie text-to-video base model without additional inference-time processing, our approach (implemented with LaVie as the base model) demonstrates a consistently higher score on all four EvalCrafter metrics, showing general and direct performance lifts across multiple facets of video generation output quality contributed by the addition of our inference-time approach. This result would be consistent with the assumption that increasing compute at inference time can be used to improve output quality with an unchanged base model. We report the qualitative performance of 3R in Appendix \ref{app:qualitative}. Fig. \ref{fig:qual_semantic_and_fiction} and Fig. \ref{fig:qual_text} demonstrate 3R's superior text-video alignment performance in complex prompts such as mushroom growing out of a human head. 3R can also visualize fictional characters such as Pikachu Jedi and understand the meaning of close-up or zoom-in better.

\begin{table}[!t]
\caption{Results on EvalCrafter benchmark. The \colorbox{first}{\textbf{first}} and \colorbox{second}{second} best results in each column are highlighted in the corresponding colors. 3R achieves the best total result, and either best or second best results in most of the individual metrics.}
\vspace{\baselineskip}
\label{tab:evalcrafter}
\begin{center}
\begin{tabular}{lccccc}
\hline
Model & Total Score & \makecell{Motion \\Quality} & \makecell{Text-Video \\ Alignment} & \makecell{Visual \\Quality} & \makecell{Temporal \\Consistency} \\ 
\hline
Show-1 & 229 & 53.74 & 62.07 & 52.19 & 60.83\\
LaVie & 234 & 52.83 & \cellcolor{second}68.49 & 57.99 & 54.23 \\ 
IPO & 234 & 53.39 & 54.62 & \cellcolor{second}62.56 & \cellcolor{first}\textbf{63.40}\\
Videocrafter2 & \cellcolor{second}243 & \cellcolor{first}\textbf{54.82} & 63.16 & \cellcolor{first}\textbf{63.98} & 61.46\\
3R & \cellcolor{first}\textbf{245} & \cellcolor{second}54.72 & \cellcolor{first}\textbf{68.73} & 58.79 & \cellcolor{second}62.65\\
\hline
\end{tabular}
\end{center}
\end{table}


\textbf{Importance of the Initial Prompt Augmentation.}  
As shown in Table~\ref{tab:ablation} (row 2), incorporating RAG-based prompt augmentation significantly improves performance. The total score increases by +7, motion quality improves by +2, and temporal consistency improves by +4, with only a slight decrease of 1 point in text-video alignment. These results underscore the importance of high quality initial prompts, suggesting that the limitations of the base model are often rooted in under-specification in user prompts rather than architectural incapacity. 

\textbf{Effectiveness of Increasing Test-Time Compute.}
Our results in Table \ref{tab:ablation} demonstrate that increasing test-time compute is a highly effective training-free strategy to close the performance gap between base models and state-of-the-art video generators. By shifting the burden from model parameters to inference-time logic, specifically through LLM-based prompt refinement, multiple-candidate sampling for video selection, and temporal interpolation, we observed a cumulative increase in the total score from 234 to 245. We report the details of other experiments in Appendix \ref{app:ablation}.

\section{Conclusion}
In this paper, we propose 3R, a novel framework for prompt optimization to improve the quality of T2V generated videos. We show that inference-time augmentation of a text-to-video model with retrieval, refinement, and ranking elements leads to performance gains in aggregate scores combining video generation metrics like motion quality, text-video alignment, visual quality, and temporal consistency. Our results contribute to a better understanding of the different inference-time pathways for improving output quality when using text-to-video models in a black-box setting.

Despite these gains, the 3R pipeline introduces increased inference latency due to multiple-candidate sampling and dense temporal interpolation. Furthermore, our ablation study highlights a critical ``feedback bottleneck": contemporary vision-language models (VLMs) often provide over-corrective or semantically drifted critiques. Future research will explore more efficient sampling strategies and video-critique architectures that provide more grounded feedback, potentially enabling a truly iterative ``generate-and-verify" loop that avoids the pitfalls of current VLM over-correction. 


\clearpage
\bibliography{iclr2026_conference}
\bibliographystyle{iclr2026_conference}
\clearpage

\appendix

\section{3R Model}
\subsection{Algorithm}
\label{app:algorithm}
\begin{algorithm}
\caption{3R Methodology}
\begin{algorithmic}[1]
\Procedure{GenerateVideo}{$I, \mathcal{D}, N$} \Comment{$I$: User Intent, $\mathcal{D}$: Database, $N$: Candidates}
    
    \State \textcolor{blue}{/* Step 1: Retrieval */}
    \State $\bm{e}_I \gets \phi(I)$ \Comment{Encode user intent using embedding function $\phi$}
    \State $P_{ret} \{ (p_j \in \mathcal{D} \mid \text{cosine\_similarity}(\bm{e}_I, \phi(p_j)) > \tau \}$
    
    \State \textcolor{blue}{/* Step 2: Refinement \& Merging */}
    \State $M_{LLM} \gets \text{LLM\_Reasoning}(I, P_{ret})$ \Comment{Merge user intent with retrieved knowledge}
    \State $\{P_1, \dots, P_N\} \gets \text{GenerateVariants}(R_{LLM}, N)$ \Comment{Sample $N$ refined prompts}
    
    \State \textcolor{blue}{/* Step 3: Generation */}
    \For{$n \gets 1$ \textbf{to} $N$}
        \State $V_n \gets \mathcal{G}(P_n)$ \Comment{Generate candidate video using T2V model $\mathcal{G}$}
    \EndFor
    
    \State \textcolor{blue}{/* Step 4: Ranking */}
    \For{$n \gets 1$ \textbf{to} $N$}
        \State $\text{TotalScore}_n \gets 0$
        \For{$i \gets 1$ \textbf{to} 29} \Comment{Evaluate 29 weighted VQA questions}
            \State $s_{i,n} \gets f_{vqa}(V_n, Q_i)$
            \State $\text{TotalScore}_n \gets \text{TotalScore}_n + (\omega_i \times s_{i,n})$
        \EndFor
    \EndFor
    \State $V^* \gets V_{\arg\max}(\text{TotalScore})$ \Comment{Select best candidate}
    
    \State \textcolor{blue}{/* Step 5: Enhancement */}
    \State $V_{final} \gets \mathcal{E}(V^*)$ \Comment{Apply super-resolution/smoothing $\mathcal{E}$}
    
    \State \Return $V_{final}$
\EndProcedure
\end{algorithmic}
\end{algorithm}

\subsection{3R Method}
\label{app:method}
\paragraph{Video Selection}
We adopt a video selection model $f_{vqa}$ that evaluates each generated video by asking a set of questions $Q_i$ covering text-video alignment, motion smoothness, and visual quality. Each question is associated with a learned weight $w_i$ that reflects how strongly it correlates with human video preferences. Some highly weighted questions are illustrated in Table \ref{tab:top5weights}. The highest weights are assigned to questions related to prompt alignment (e.g., whether the video satisfies all requirements of the text), physical realism (e.g., whether object motion is realistic), and fine detail quality. In contrast, questions pertaining to subjective aesthetics (e.g., whether lighting is beautiful) receive much smaller weights. Consequently, the reward model places greater emphasis on semantic correctness and physical plausibility, allowing it to reliably select the best candidate among a set of generated videos.
\begin{equation}
    V^* = \arg\max_{V_n} \mathcal{S}(V_n) \quad \text{where} \quad \mathcal{S}(V_n) = \sum_{i=1}^{29} w_i \cdot f_{vqa}(V_n, Q_i)
\end{equation}

\begin{table}[!t]
\caption{Top 5 weighted questions used in VisionReward-Video scoring  out of 29 questions. Higher scores are assigned to text-video alignment questions.}
\vspace{\baselineskip}
\label{tab:top5weights}
\centering
\begin{tabular}{p{0.07\linewidth} p{0.12\linewidth} p{0.72\linewidth}}
\toprule
\textbf{Rank} & \textbf{Weight} & \textbf{Question} \\
\midrule
1 & 1.1418 &
Does the video not completely fail to meet the requirements stated in the text ``[prompt]''? \\
2 & 0.9544 &
Does the video meet all the requirements stated in the text ``[prompt]''? \\
3 & 0.4390 &
Is the object's movement completely realistic? \\
4 & 0.4293 &
Are the details very refined? \\
5 & 0.3942 &
Is the video content part of the physical world? \\
\bottomrule
\end{tabular}
\end{table}

\paragraph{Video Enhancement}
Previous work \citep{temp_inter_dynamic,temp_inter_inpainting} has used temporal interpolation techniques to learn meaningful representations for dynamic scenes. In our approach, our goal is to specifically improve motion smoothness and enrich temporal detail by extracting features through a temporal interpolation network $\mathcal{E}$. This network first duplicates current frames to match the number of target frames with randomly initialized gaussian noise. Then it uses a pre-trained UNet diffusion model \citep{ronneberger2015u} to denoise and results in interpolated frames. Besides, it uses the user prompt to guide the interpolation process to ensure proper temporal coherence and alignment with the user intent.
\begin{equation}
    V_{final} = \mathcal{E}(V^*) 
\end{equation}

\begin{table}[!t]
    \caption{Comparison of Prompt Optimization Outputs. RAPO contains information such as `eating hay'  or `shooting competition' that has no relevance to the user intent `a tranquil tableau of barn' while 3R filters out irrelevant modifiers and generates relevant ones like `wooden structure' and `peaceful countryside setting'.}
    \vspace{\baselineskip}
    \label{tab:merge}
    \centering
    \begin{tabular}{|l|p{0.8\linewidth}|} 
    \hline
    \textbf{Category} & \textbf{Content} \\
    \hline
    \textbf{User Prompt} & A tranquil tableau of barn. \\
    \hline
    \makecell[l]{\textbf{Retrieved} \\ \textbf{Modifiers}} & a lively and energetic performance, walking inside, eating hay, being kept in the barn for some purpose, appears to be a farmer, a lively and joyful scene, wearing pink outfits, in a barn, preparing for a shooting competition, dancing in a barn, laying on the ground, a barn, opening the door of a barn, inside a barn \\
    \hline
    \textbf{RAPO} & A lively and energetic performance is taking place inside a barn. A farmer, wearing pink outfits, is walking inside, eating hay, and appears to be preparing for a shooting competition. The scene is lively and joyful, and the animals seem to be laying on the ground. The farmer opens the door of the barn and invites the audience to witness the performance, which includes dancing in the barn. \\
    \hline
    \textbf{3R} & In a peaceful countryside setting, a barn sits quietly surrounded by lush green fields. The wooden structure, weathered yet sturdy, is bathed in the warm glow of the afternoon sun.\\
    \hline
    \end{tabular}
\end{table}

\section{Experiments}
\subsection{Implementation Details}
\label{app:implementation}
We use the relation graph from RAPO \citep{Gao_2025_CVPR} to extract relevant and useful modifiers. As the sentence transformer, we use all-MiniLML6-v2 to get the embeddings of sentences \citep{wang2020minilm}. To merge the retrieved modifiers with the user prompt, we use Mistral model \citep{jiang2023mistral7b} with cosine similarity threshold $\tau=0.5$ between an user prompt and a modifier. In our final model, we use GPT4o \citep{openai2024gpt4technicalreport} as the prompt refiner. We create 4 prompt candidates with variations by keeping the original user intent intact. As our final base text2video generation model, we use Lavie \citep{wang2023lavie}. We choose Lavie because it is faster than other diffusion models like Wan \citep{wan2025wan} and generates quality video using minimal resources. In our Nvidia H200 GPU, it takes around 5s to generate one video. As for the video selection model, we use Vision-Reward Model \citep{vision_reward}. It generates scores for all 4 candidate videos using multimodal visual question answering technique. The selected video is used as input to the temporal interpolation network \citep{wang2023lavie} that increases the number of frames to 61. For the VLM Critique ablation study, we use GPT4o \citep{openai2024gpt4technicalreport}. We run the overall pipeline two times with two random seeds and report the average results in Table \ref{tab:evalcrafter}.
\vspace{2cm}

\subsection{Qualitative Results}
\label{app:qualitative}
In this section, we illustrate a few challenging examples from the EvalCrafter benchmark. We compare the qualitative performance of 3R with Lavie base model and IPO.
\begin{figure}[!h]
    \centering
    \includegraphics[width=\linewidth, trim=40 235 40 45, clip]{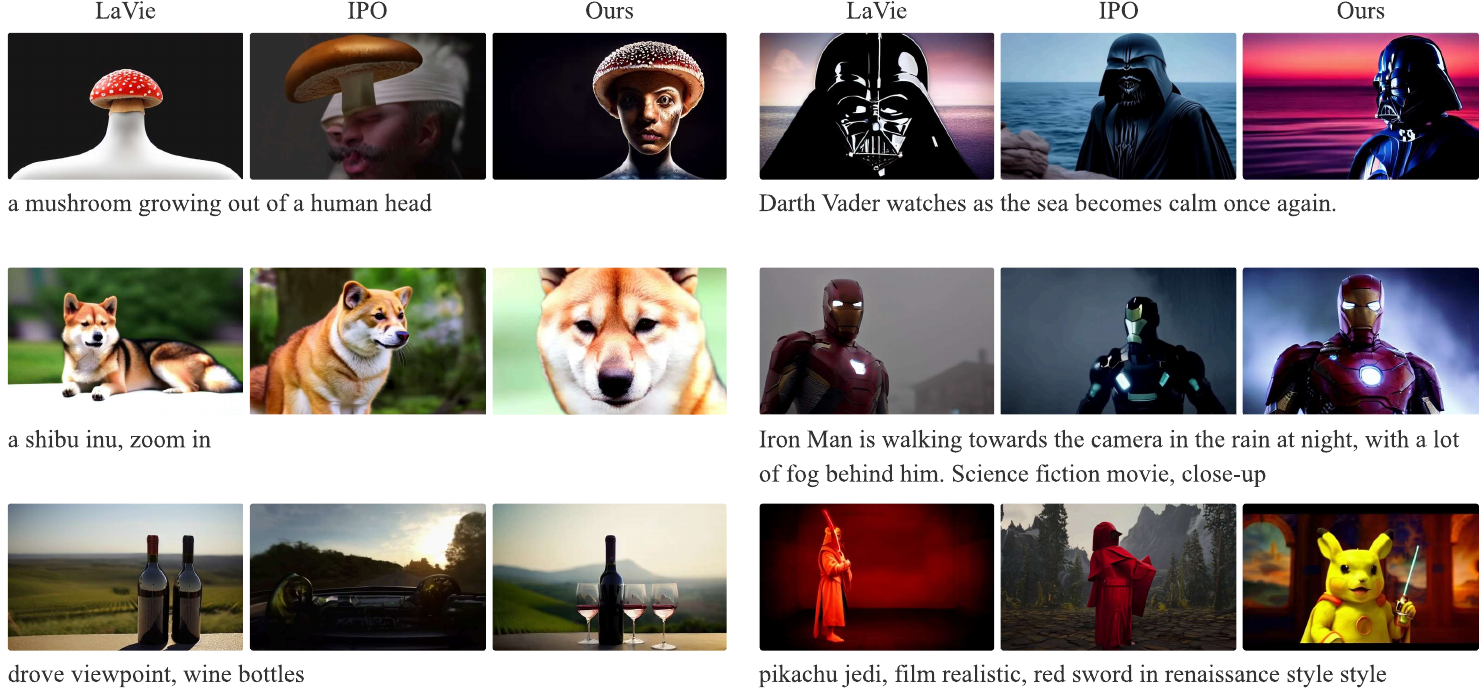}
    \caption{\textbf{Qualitative comparison of Lavie, IPO, and 3R in two common video generation failure modes.} The left side shows prompts and video frames representing challenges in semantic alignment such as mushroom growing out of human head or zoom-in and the right side shows prompts and video frames representing challenges in addressing fictional references such as Darth Vedar or Pikachu Jedi.
    }
    \label{fig:qual_semantic_and_fiction}
\end{figure}

\begin{figure}[!t]
    \centering
    \includegraphics[width=0.5\linewidth]{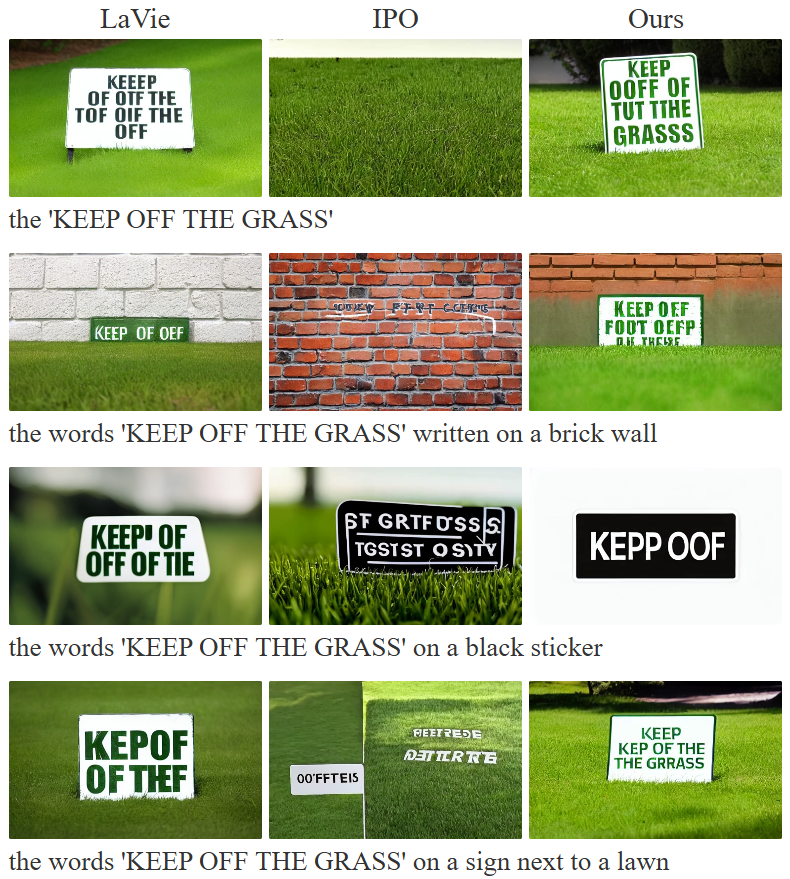}
    \caption{\textbf{Qualitative comparison of approaches in the common failure mode of generating videos containing text.} 
    We compare the first frame of the videos generated by Lavie (left), IPO (middle), and 3R (right) in the common failure mode of text generation in videos, as observed from prompts provided by the EvalCrafter benchmark. All three approaches show strong limitations in generating correct text, but 3R manages to generate qualitatively more legible text where the intended text in the prompt ("keep off the grass" or "keep off") can still be partially inferred despite typos. The prompts and respective video frames show how our approach can address prompts requiring multiple semantic conditions while producing less distorted outputs.
    }
    \label{fig:qual_text}
\end{figure}

\subsection{Ablation Study}
\label{app:ablation}
We explain the results of other research questions in details here.

\textbf{Impact of Video Selection.}  
Table~\ref{tab:ablation} (row 3) presents the impact of incorporating the video selection module. Adding a vision-based reward through diffusion-based human preference alignment increases the total score by +2, driven primarily by a +2 improvement in text-video alignment. This highlights the crucial role of preference alignment. The selection module reliably filters out semantically inconsistent generations. Example videos show that the selected outputs more faithfully reflect user intent compared to their unfiltered counterparts.

\textbf{Temporal Interpolation and Consistency.}  
Increasing the number of frames generated from 16 to 61 leads to a noticeable improvement in temporal smoothness, as reflected in the temporal consistency score. Videos with higher frame density exhibit reduced flicker, smoother motion trajectories, and fewer disjoint transitions. Offline visual comparisons clearly show the improvement in motion coherence, particularly in scenes with significant camera or object movement.

\textbf{Efficacy of VLM Critque}  
As shown in Table~\ref{tab:ablation} (row 4), introducing a video critique module does not produce measurable performance gains. Inspection of the VLM-generated critique text reveals several misleading or incorrect interpretations, frequently exhibit semantic drift or unnecessary over-corrections, underscoring the unreliability of critique signals for this task. Two examples are illustrated in Appendix \ref{app:vlm_critque} where VLM either tries to over-correct or the T2V model fails to follow the VLM instructions, resulting in degradation in video quality in both cases. The specific prompt used to extract critique data is detailed in Appendix \ref{app:vlm}.

Overall, the ablation results confirm that each component: RAG-based augmentation, diffusion-based preference alignment, and temporal-aware interpolation, contributes meaningfully to the 3R pipeline, offering complementary improvements across evaluation dimensions.

\begin{table}[h]
\caption{Ablation results. The \colorbox{first}{\textbf{first}} and \colorbox{second}{second} best results in each column are highlighted in the corresponding colors. Initial prompt refinement, video selection model, and temporal interpolation model each contribute to the total score while vision-language model critique degrades text-video alignment severely due to it's over-correction nature.}
\vspace{\baselineskip}
\label{tab:ablation}
\begin{center}
\begin{tabular}{lccccc}
\hline
Model & Total Score & \makecell{Motion \\Quality} & \makecell{Text-Video \\ Alignment} & \makecell{Visual \\Quality} & \makecell{Temporal \\Consistency} \\ 
\hline
LaVie (Baseline) & 234 & 52.83 & 68.49 & 57.99 & 54.23 \\ 
One Prompt & 241 & \cellcolor{second}54.73 & 67.54 & 59.75 & 58.75\\
$N$ Prompts + Video Selection & \cellcolor{second}243 & 54.64 & \cellcolor{first}\textbf{69.44} & \cellcolor{second}59.86 & 58.94\\
Video Selection + VLM Critique & 241 & \cellcolor{first}\textbf{54.89} & 66.35 & \cellcolor{first}\textbf{60.20} & \cellcolor{second}59.74\\
Video Selection + Temporal Inter. & \cellcolor{first}\textbf{245} & 54.72 & \cellcolor{second}68.73 & 58.79 & \cellcolor{first}\textbf{62.65} \\ 
\hline
\end{tabular}
\end{center}
\end{table}

\newpage
\section{Prompts}
In this section, we report all LLM prompts in the same format we used in this study.
\subsection{User Prompt Refinement}
\label{app:refinement}
\begin{tcolorbox}[breakable]
\begin{verbatim}
system_prompt = f"""
You are a precise and creative language model specialized in
refining scene descriptions for text-to-video generation.

Your goal:
1. **Preserve creative imagination**, including surreal or 
impossible but visually interesting elements (e.g., "glass 
tree”, "cat and fish swimming together”).
2. **Remove or rewrite illogical, impossible, or 
linguistically nonsensical phrases** unless user prompt is
asking for it. (e.g., "window is singing”, "barn is well-
behaved”, "sea is dry”, "moon shines on the sun”, "a toy is
playing itself").
3. **Ensure completeness by checking missing keywords:**
Extract key subjects, objects, and actions from the user
prompt, and check if they are present and relevant in the
description. If they are missing in the description, add 
them meaningfully to maintain semantic alignment. (e.g. user
prompt is "a boat and a fish" but description does not 
contain any info about fish.)
4. **Extract characters**: Extracts all characters and 
counts from the user prompt. (e.g. 3 persons, 2 cats, 5
birds, one dog)
5. **Preserve user prompt characters**: Ensure the 
subject/object/character extracted above are included in the
refined description. If the description is missing one or 
more characters, add them meaningfully. (e.g, user prompt 
mentions 3 persons but descriptions mention 1 person)
6. **Preserve colors** Ensure each character color is 
preserved in the final prompt (e.g red apple, yellow tree)
7. **Preserve Video Style**: Ensure video style adheres to
user prompt. (e.g if user prompt does not say "animation",
description must not mention "animation" style video.
8. Maintain temporal coherence and remove unnecessary
repetition or contradictions.
9. Produce 4 clean, vivid, logically consistent scene 
description candidates suitable for text-to-video models,
word limit: 100 words.
10. Make necessary variations among the 4 descriptions but
you MUST ensure they all adhere to above rules.

Return only the 4 **refined descriptions** in English
without commentary as **a single list of 4 strings** without 
any numbering.
"""

# Few-shot examples for reference
examples = f"""
Example 1:
User Prompt: A mountain cabin during snowfall
Original Description:
A small cabin sits on the mountain while snow falls gently.
The cabin is burning in the snow, and inside it’s raining 
heavily. The snow is hot, and the fireplace is filled with
ice. The cabin is empty but also full of people singing.
Refined Description:
A small wooden cabin rests quietly on a snowy mountain 
slope. Snowflakes drift through the air, and warm light
glows from the windows, creating a peaceful and cozy winter
atmosphere.            
Example 3:
User Prompt: A city skyline at sunset
Original Description:
The city skyline at sunset is filled with colorful buildings
that change colors every second. The river under it is above
the buildings, creating a surreal view.
Refined Description:
The city skyline glows in warm shades of orange and pink as
the sun sets. The river below reflects the tall buildings, 
creating a calm and beautiful evening scene.
Example 5:
User Prompt: A shark and a cat
Original Description:
A shark is swimming in an ocean, looking for food. Clear 
blue water looks beautiful while the shark is swimming. The
sun is shining brightly.
Refined Description:
A shark is swimming in shallow water, while a cat is walking
on the nearby beach. It is a sunny day, and the blue water 
makes a beautiful scene.
Example 6:
User Prompt: A person is walking with two dogs in a park.
Original Description: 
A person is walking with his dog in a park. The dog is 
playing around. The park has green grass and some gree trees
nearby. Overall scene is peaceful.
Refined Description:
A person is walking with two dogs in a park full of small 
green grass. The dogs are playing with each other and walking
beside the person. There are some trees nearby. Overall a 
beautiful afternoon for a refreshing walk.
"""

# Combine examples with user data
final_prompt = f"""
{examples}

Now refine the following description according to the above
rules and examples. Keep it logically coherent, concise, and
visually descriptive.

User Prompt: {user_prompt}
Original Description: {description}

4 Refined Descriptions:
"""
\end{verbatim} 
\end{tcolorbox}

\subsection{Vision Language Model Feedback Prompt}
\label{app:vlm}
\begin{tcolorbox}[breakable]
\begin{verbatim}
You are an **Expert Text-to-Video (T2V) Alignment and 
Optimization Agent**. Your function is to critically analyze
the generated video against two distinct prompts:
1.  **User Prompt Intent (UPI):** The short, original user 
instruction (the truth source).
2.  **Description Prompt Old (DPO):** The detailed prompt
that was actually fed to the T2V model.

Your task is to prioritize Fidelity to UPI and strictly output
a single, valid JSON object following the prescribed schema.
Your analysis must use multi-step reasoning (Chain-of-Thought)
to link video failure to DPO flaws.

**Inputs for Analysis:**
1.  **User Prompt Intent (UPI):**"{USER_PROMPT_INTENT}"
2.  **Description Prompt Old (DPO):**"{DESCRIPTION_PROMPT_OLD}"

**PHASE 1: MULTI-DIMENSIONAL ASSESSMENT**
Evaluate the video on a scale of 0 to 10. For each dimension, 
record the most critical observation (1-2 sentences).
* **A_TV (Text-Visual Alignment):** Adherence to all objects,
attributes like color, count, and environment in the DPO and UPI.
* **C_T (Temporal Coherence):** Consistency of identity, 
background, and motion quality across the video duration.
* **F_C (Compositionality Fidelity):** Accuracy of complex 
relations (spatial, numerical, causal).
* **Q_V (Visual Quality):** Resolution, aesthetic, and 
freedom from artifacts.

**PHASE 2: ROOT CAUSE DIAGNOSIS (Chain-of-Thought)**
[STEP 2.1: IDENTIFY BOTTLENECK] Select the lowest-scoring
dimension. Prioritize A_TV, C_T, or F_C over Q_V.
[STEP 2.2: ANALYZE DUAL-PROMPT ALIGNMENT] Did the DPO (and 
subsequently the video) violate the UPI?
[STEP 2.3: FORMULATE DIAGNOSIS] Write a concise 'root_cause'
linking the flaw (e.g., Compositional Overload, Ambiguous 
Temporal Dynamics) to a specific part of the DPO.

**PHASE 3: ACTIONABLE PROMPT PRESCRIPTION**
Generate the refined prompts based solely on the Phase 2 
Diagnosis.
* **P_new (Description Prompt New) Generation Rules:** Rewrite 
the DPO to fix the flaw and enforce UPI alignment. Use
explicit staging, token weighting (e.g., (KEYWORD:1.3)), and
structural formatting.

**STRICT JSON OUTPUT SCHEMA (Start your response here):**

```json
{{
  "prompt_analysis": {{
    "user_prompt_intent": "{USER_PROMPT_INTENT}",
    "bottleneck_flaw_metrics": {{
      "metric": "A_TV" | "C_T" | "F_C" | "Q_V",
      "score": 0,
      "observation": "..."
    }},
    "diagnosis": {{
      "root_cause": "..."
    }}
  }},
  "prescription": {{
    "P_new": "...",
    "meta_instruction": "..."
  }}
}}
\end{verbatim}
\end{tcolorbox}

\subsection{VLM Feedback Output}
\label{app:vlm_critque}

\begin{tcolorbox}[title=VLM proposes but T2V fails to follow, breakable]
\begin{verbatim}
"user_prompt": "Teddy bear and 3 real bear",
"Iterations_1": {
    "description_prompt_old": "Teddy bear and a real bear are
    playing together, with the teddy bear being a stuffed toy,
    while the real bear is a live animal.",
    "metrics": {
      "metric": "A_TV",
      "score": 2,
      "observation": "The video depicts only teddy bears without 
      any real bears, failing to align with the UPI."
    },
    "prompt_new": "A forest clearing with a single teddy bear 
    and three real bears. The teddy bear (STUFFED TOY:1.3) sits 
    in front of the (REAL BEARS:1.5). The real bears are natural 
    and curious, engaging with the environment."
  },
"Iterations_2": {
    "description_prompt_old": "A forest clearing with a single
    teddy bear and three real bears. The teddy bear 
    (STUFFED TOY:1.3) sits in front of the (REAL BEARS:1.5). 
    The real bears are natural and curious, engaging with the 
    environment.",
    "metrics": {
      "metric": "A_TV",
      "score": 0,
      "observation": "The video displays three stuffed teddy
      bears consistently, and no real bears are present, 
      violating both the UPI and DPO."
    },
    "prompt_new": "A forest clearing with a single teddy bear 
    (STUFFED TOY:1.3) juxtaposed with three real bears (LIVE 
    ANIMALS:1.5). The teddy bear sits distinctly in front of the 
    real bears, which are naturally curious and engaging with 
    the environment."
  },    
\end{verbatim}
\end{tcolorbox}

\begin{tcolorbox}[title=VLM worsens quality]
\begin{verbatim}
"user_prompt": "A small bird sits atop a blooming flower stem.",
"Iterations_1": {
    "description_prompt_old": "A small bird perches delicately
    atop a blooming flower stem. The vibrant petals surround
    the bird, creating a picturesque scene of nature's beauty.",
    "metrics": {
      "metric": "A_TV",
      "score": 6,
      "observation": "The video partially aligns with the UPI as
      it depicts a bird on a flower stem but misses the 
      delicacy and integration with the bloom that the prompt
      suggests."
    },
    "prompt_new": "A small bird (KEYWORD:1.3) sits (ACTION:1.2)
    atop a blooming flower stem. Focus on the vibrancy of the
    flower and the bird's delicate placement. 
    Ensure the petals and stem are clearly shown."
  },
"Iterations_2": {
    "description_prompt_old": "A small bird (KEYWORD:1.3) sits
    (ACTION:1.2) atop a blooming flower stem. Focus on the 
    vibrancy of the flower and the bird's delicate placement. 
    Ensure the petals and stem are clearly shown.",
    "metrics": {
      "metric": "A_TV",
      "score": 2,
      "observation": "The video focuses only on the flower and
      does not include the bird."
    },
    "prompt_new": "A small bird (KEYWORD:1.5) clearly sits
    (ACTION:1.3) atop a blooming flower stem, ensuring both the 
    bird and flower are fully visible."
  }
\end{verbatim}
\end{tcolorbox}

\end{document}

%% file: math_commands.tex
\usepackage{amsmath,amsfonts,bm}

\def\eqref#1{equation~\ref{#1}}

\def\1{\bm{1}}

\DeclareMathAlphabet{\mathsfit}{\encodingdefault}{\sfdefault}{m}{sl}
\SetMathAlphabet{\mathsfit}{bold}{\encodingdefault}{\sfdefault}{bx}{n}

